\documentclass{article}

\PassOptionsToPackage{numbers, compress}{natbib}
\usepackage[final]{nips_2018}

\usepackage[utf8]{inputenc} % allow utf-8 input
\usepackage[T1]{fontenc}    % use 8-bit T1 fonts
\usepackage{hyperref}       % hyperlinks
\usepackage{url}            % simple URL typesetting
\usepackage{booktabs}       % professional-quality tables
\usepackage{amsfonts}       % blackboard math symbols
\usepackage{nicefrac}       % compact symbols for 1/2, etc.
\usepackage{microtype}      % microtypography

\usepackage[newfloat=true,frozencache=true]{minted}
\usepackage{xcolor}

\usepackage{subcaption}
\usepackage{pgfplots}
\pgfplotsset{compat=1.14}
\usepackage{filecontents}

\title{Tangent: Automatic differentiation using source-code transformation for dynamically typed array programming}

\author{
  Bart van Merri\"enboer \\
  MILA, Google Brain \\
  \texttt{bartvm@google.com} \\
  \And
  Dan Moldovan \\
  Google Brain \\
  \texttt{mdan@google.com} \\
  \And
  Alexander B Wiltschko \\
  Google Brain \\
  \texttt{alexbw@google.com} \\
}

\begin{document}

\maketitle

\begin{abstract}

The need to efficiently calculate first- and higher-order derivatives of increasingly complex models expressed in Python has stressed or exceeded the capabilities of available tools. In this work, we explore techniques from the field of \emph{automatic differentiation} (AD) that can give researchers expressive power, performance and strong usability. These include source-code transformation (SCT), flexible gradient surgery, efficient in-place array operations, higher-order derivatives as well as mixing of forward and reverse mode AD. We implement and demonstrate these ideas in the Tangent software library for Python, the first AD framework for a dynamic language that uses SCT.

\end{abstract}

\section{Introduction}

  Many applications in machine learning rely on gradient-based optimization, or at least the efficient calculation of derivatives of models expressed as computer programs. Researchers have a wide variety of tools from which they can choose, particularly if they are using the Python language \cite{paszke2017automatic, maclaurin2015autograd, tokui2015chainer, al2016theano, abadi2016tensorflow}. These tools can generally be characterized as trading off \emph{research} or \emph{production} use cases, and can be divided along these lines by whether they implement automatic differentiation using operator overloading (OO) or SCT. SCT affords more opportunities for whole-program optimization, while OO makes it easier to support convenient syntax in Python, like data-dependent control flow, or advanced features such as custom partial derivatives. We show here that it is possible to offer the programming flexibility usually thought to be exclusive to OO-based tools in an SCT framework.
  
  Tangent is the first AD framework using SCT in a dynamically typed language. We produce efficient derivatives using a novel combination of multiple dispatch, lazy evaluation, and static optimizations. Further, Tangent has mutable multidimensional arrays as first class objects, implemented using persistent data structures for performance in the context of reverse mode AD.\@ By operating directly on Python source code, Tangent is able to achieve a separation of concerns that other AD libraries do not. Specifically, we achieve compositionality with tools in the Python ecosystem, such as debuggers, profilers and other compilers. Tangent makes it easy and efficient to express machine learning models, and is open source \footnote{Source code and documentation available at \url{https://www.github.com/anonymous}}.

\section{Background}

Automatic differentiation (AD) is a set of techniques to evaluate derivatives of mathematical functions defined as programs~\cite{griewank2008evaluating}, and is heavily used in machine learning~\cite{baydin2015automatic}. It is based on the insight that the chain rule can be applied to the elementary arithmetic operations (primitives) performed by the program. This allows derivatives to be calculated up to machine precision~\cite{naumann2012art} with only a constant overhead per operation. AD is different from symbolic differentiation (which applies to mathematical expressions instead of programs) and numerical differentiation (where the gradient is approximated using finite differences).

For multidimensional functions, \(f: \mathbb{R}^n \rightarrow \mathbb{R}^m\), the application of the chain rule results in a series of matrix-vector multiplications involving the primitives' Jacobians and partial derivatives of intermediate values. The order in which these multiplications are evaluated determines the runtime complexity. Forward-mode AD evaluates the chain rule from inside to outside and is efficient for functions where \(m > n\). The implementation of forward mode is relatively straightforward, since the partial derivatives are evaluated in step with the primitives. Forward mode is commonly implemented by replacing numbers with \emph{dual numbers}, which can be interpreted as a variable's value along with its partial derivative with respect to one of the inputs. Reverse-mode AD, where the chain rule is evaluated from outside to inside, is more efficient in the case where \(n > m\). Reverse mode is more complex to implement because evaluation of the partial derivatives requires reversing the execution order of the original program. This reversal gives rise to a non-local program transformation where the beginning of the original program interacts with the generated derivative program.

Two methods of implementing reverse-mode AD are commonly distinguished: operator overloading (OO) and source code transformation (SCT). In the OO approach primitives are overloaded so that at runtime each numerical operation is logged onto a tape (a linear trace) along with its inputs. The chain rule can then be evaluated by walking this tape backward. SCT, on the other hand, explicitly transforms the original program (primal) prior to execution to produce a separate derivative function (adjoint) whose control flow is the reverse of the original program. Both approaches have different implementation, performance, and usability trade-offs~\cite{bischof2000computing}.

OO is easier to implement and since it only requires tracing, it naturally supports all the features of the host language such as higher-order functions, recursion, and classes. If the control flow of the program is data dependent, the function must be retraced for each function call, which can cause significant overhead when the runtime of the primitives is small compared to the cost of tracing. Since the adjoint program is run by a separate `derivative interpreter' (the algorithm that walks the tape in reverse), there is no adjoint program that can be inspected, optimized or compiled.

SCT is harder to implement, since it requires tooling to transform intermediate representations of computer programs. Further, the AD tool must explicitly support all of the features of the host language, including function calls, loops, classes, etc. If a language feature is not explicitly handled by the AD system, the user cannot take derivatives of code using those features. For some languages like C and C++ this requires a separate toolchain, but reflective languages such as Lisp and Python contain the necessary tools to capture, transform, and output program representations. The advantage of SCT is that there is no runtime overhead, and that generated derivative code can be statically analyzed and optimized.

\section{Prior work}

AD packages using either approach have long existed for, e.g., C, C++, Fortran, (see~\cite{baydin2015automatic} for an overview) and have been used in fields such as computational fluid dynamics, atmospheric sciences, and astronomy. In the machine learning community different needs have led to the development of a separate set of tools. In particular, the community has a strong attachment to Python and its models rely heavily on multidimensional arrays.

Theano~\cite{al2016theano} and TensorFlow~\cite{abadi2016tensorflow} are two popular machine learning frameworks with support for SCT AD\@. Although Python-based, they do not perform AD on the Python code. Instead, Python is used as a metaprogramming language to define a dataflow graph (computation graph) on which SCT is performed. Since these dataflow graphs only operate on immutable values and do not have function calls or lexical scoping, the AD logic is simplified. The same graph representation is then used for static analysis, optimizations, and code generation.

OO has been used to implement AD in Python in packages such as Autograd~\cite{maclaurin2015autograd}, Chainer~\cite{tokui2015chainer}, and PyTorch~\cite{paszke2017automatic}.

Although OO frameworks are easier to implement, their runtime performance falls short of that of frameworks using SCT for workloads that do not spend most of their time in hand-optimized compute primitives.\@ On the other hand, existing frameworks that use SCT require the user to metaprogram computation graphs, significantly complicating the definition of ML models. Tangent applies SCT directly on the Python language in order to combine the performance achieved by SCT with the usability of programming directly in Python.

% TODO: more detail on Chainer/PyTorch vs Autograd?
% TODO: emphasize that AD always applies to a particular program representation, and that is what primarily varies across different implementation stratgies?

  \section{Features}
  
  Tangent supports reverse mode and forward mode, as well as function calls, loops, and conditionals. Higher-order derivatives are supported, and reverse and forward mode can readily be mixed. To our knowledge, Tangent is the first SCT-based AD system for Python and moreover, it is the first SCT-based AD system for a dynamically typed language. As a consequence of performing SCT directly on the Python source code, the generated programs can be run, inspected, profiled, and debugged with standard Python tools. Tangent supports array programming on both CPU and GPU through the NumPy~\cite{oliphant2006guide} and TensorFlow Eager libraries. A modular design makes it possible to extend Tangent to support other numeric libraries.
  
  The ability to write code directly in Python makes Tangent less verbose and more idiomatic than the metaprogramming approach used by Theano and Tensorflow (see Listing~\ref{lst:metaprogramming}). Moreover, the metaprogrammed code requires a separate compiler and/or runtime, separate debugging tools, etc.
  
  \begin{listing}[ht]
    \begin{minipage}[t]{0.65\textwidth}
    \begin{minted}[fontsize=\small]{python}
x = tf.placeholder(tf.float32)
y = x * x
dx, = tf.gradients(y, x)

with tf.Session() as sess:
    dx_ = sess.run(dx, feed_dict={x: 3})
    \end{minted}
    \vspace{5pt}
    \end{minipage}
    \begin{minipage}[t]{0.35\textwidth}
    \begin{minted}[fontsize=\small]{python}
def f(x):
    return x * x

df = grad(f)
dx = df(3)
    \end{minted}
    \end{minipage}
    \begin{minipage}[t]{0.49\textwidth}\subcaption{TensorFlow requires the programmer to define the variable \texttt{x} as part of the dataflow graph. After the program (dataflow graph) has been constructed, its evaluation must be triggered by creating a session and providing values for the arguments.}    \label{lst:metaprogramming}
\end{minipage}
    \hfill
    \begin{minipage}[t]{0.49\textwidth}\subcaption{Tangent and libraries such as Autograd allow the user to write pure Python.}\end{minipage}
    \caption{Comparison between metaprogramming and direct programming approaches.}
\end{listing}
  
The OO approach can be problematic for debugging and usability as well as performance (see Listing~\ref{lst:overhead}). When an adjoint function \mintinline{python}{grad(f)} is called, the function \mintinline{python}{f} is executed with non-standard semantics, since each function and operator has been overloaded to log onto a tape, after which the tape is walked in reverse using a loop that is internal to the framework. This means that each function call incurs tracing overhead, and errors that occur during execution will potentially have tracebacks involving tracing logic that can be hard for a user to decipher.

  \begin{listing}[ht]
  \centering
  \begin{minipage}[t]{0.4\textwidth}
    \begin{minted}[xleftmargin=0.2\textwidth,fontsize=\small]{python}
def f(x):
    while x < 10000:
        x = x + 1
    return x
    \end{minted}
       \caption{In the case that \texttt{x} is a scalar, this trivial program and its derivative contain a tight loop. Since it does not require tracing, Tangent's derivative of this function is approximately 30\% faster than PyTorch's, even though PyTorch is given type information about \texttt{x} whereas Tangent's derivative is dynamically typed.}
       \label{lst:overhead}
    \end{minipage}
    \hfill
  \begin{minipage}[t]{0.55\textwidth}
    \begin{minted}[xleftmargin=0.04\textwidth,fontsize=\small]{python}
# Generated gradient function
def dfdx(x, by=1.0):
    # Grad of: y = x * x
    _bx = tangent.unbroadcast(by * x, x)
    _bx2 = tangent.unbroadcast(by * x, x)
    bx = _bx
    bx = tangent.add_grad(bx, _bx2)
    return bx
    \end{minted}
      \caption{Source code of the gradient of \mintinline{python}{def f(x): return x * x} in Tangent. The \texttt{unbroadcast} function is responsible for reversing the broadcasting performed by NumPy when performing element-wise operations on differently-sized multidimensional arrays.}
       \label{lst:grad}
    \end{minipage}
\end{listing}

The adjoint code generated by Tangent is regular Python (see Listing~\ref{lst:grad}), which means that it can be debugged using standard debuggers such as \texttt{pdb}, profiled using, e.g., \texttt{line\_profiler}, optimized by JIT compilers such as Numba and Pythran. The adjoint code can readily be inspected by users, and Tangent tries to ensure that is human-readable and commented, which is useful for debugging as well as for didactic purposes.

Unlike most existing ML frameworks, arrays in Tangent are mutable without incurring unnecessary performance loss (see Section~\ref{sec:arrays} for implementation details).

\subsection{Backward pass inlining}

Many algorithms use approximations or modifications of the gradient. For example, for performance reasons recurrent neural networks (RNNs) are often trained using truncated backpropagation through time~\cite{williams1990efficient} (TBPTT) and/or gradient clipping~\cite{pascanu2013difficulty}. In other cases, custom gradients are used to train models with discontinuous functions (e.g.\ straight-through estimators) or for many other applications ~\cite{bengio2013estimating, ganin2016domain, oord2017neural, heess2015learning, jang2016categorical, nokland2016direct, lillicrap2014random}. A user might also be interested in accessing the values of gradients for logging or debugging.

Existing AD frameworks support this functionality by allowing the user to define custom adjoints for functions. Tangent provides this functionality as well, but uses Python's context manager syntax to introduce a second, novel way of allowing the user to inject arbitrary code into the gradient computation (see Listing~\ref{lst:inject}). We believe this syntax provides a more succinct and readable way of modifying the adjoint code in many cases.

  \begin{listing}[ht]
    \begin{minipage}[t]{0.49\textwidth}
    \begin{minted}[fontsize=\small]{python}
# Original function
def f(x):
    with insert_grad_of(x) as dx:
        if dx > 10:
            print('Clipping', dx)
            dx = 10
    return x * x
    \end{minted}
    \end{minipage}
    \begin{minipage}[t]{0.49\textwidth}
    \begin{minted}[fontsize=\small]{python}
# Generated gradient function
def dfdx(x, bx_times_x=1.0):
    x_times_x = x * x
    # Grad of: dx = 10
    _bx = tangent.unbroadcast(bx_times_x * x, x)
    _bx2 = tangent.unbroadcast(bx_times_x * x, x)
    bx = _bx
    bx = tangent.add_grad(bx, _bx2)
    # Inserted code
    if bx > 10:
        print('Clipping', bx)
        bx = 10
    return bx
    \end{minted}
    \end{minipage}
       \caption{Gradient clipping implemented using Tangent. The code inside of the context manager is inserted directly into the derivative function.}
       \label{lst:inject}
       \end{listing}

\section{Implementation}

Tangent uses Python's built-in machinery to inspect and transform the abstract syntax tree (AST) of parsed source code. For each piece of supported Python syntax, we have implemented a rule indicating how to rewrite an AST node into its primal and adjoint. We have defined adjoints for e.g.\ mathematical operators, function calls to NumPy methods, and constructs such as if-statements and for-loops. The adjoints are defined using a custom template programming syntax (see Listing~\ref{lst:template}) which makes it easy for users to add new or custom derivatives.

  \begin{listing}[ht]
  \centering
  \begin{minipage}[t]{0.4\textwidth}
    \begin{minted}[xleftmargin=0\textwidth,fontsize=\small]{python}
# Templates are Python functions
@adjoint(numpy.multiply)
def adjoint_multiply(z, x, y):
    d[x] = y * d[z] 
    d[y] = x * d[z]
# If the primal contains...
c = numpy.multiply(a, b)
    \end{minted}
        \end{minipage}
        \hfill
  \begin{minipage}[t]{0.55\textwidth}
    \begin{minted}[xleftmargin=0\textwidth,fontsize=\small]{python}
# ...Tangent will expand the template...
new_ast = tangent.template.replace(
    adjoint_multiply,
    z='c', x='a', y='b')
# ...generating the following adjoint
b_a = b * b_c
b_b = a * b_c
    \end{minted}
        \end{minipage}
       \caption{Tangent's source generation uses templating. The template takes the form of a Python function which is parsed into its AST. The variable names in the AST are substituted and variables for the partial derivatives are constructed, before the AST is inserted into the code of the adjoint function.}
       \label{lst:template}
\end{listing}

Generated derivative code is constructed using the built-in Python AST. The alternative program representations are Python bytecode, which changes across Python versions, and a formatting-aware AST used in the Python 2-to-3 conversion tool, \mintinline{python}{2to3}, which has little tooling and is more cumbersome to use. We acquire and manipulate the Python AST with the \mintinline{python}{inspect} and \mintinline{python}{ast} modules from the standard library, and standardize small differences between the Python 2 and Python 3 AST with \mintinline{python}{gast} and use \mintinline{python}{astor} to invert ASTs into readable source code.

To support dynamic typing and array programming while maintaining efficiency, Tangent relies on a novel combination of multiple dispatch, lazy evaluation, persistent data structures, and static optimizations.

\subsection{Multiple dispatch}

Python is a dynamic language which uses dynamic typing, late binding and operator overloading. These fundamental features of the language make it impossible to determine ahead of time how a statement will be executed, which means it is impossible to determine ahead of time what the adjoint program should be. Instead of enforcing static types (for example by using type annotations and MyPy\footnote{\url{http://mypy-lang.org/}}), Tangent embraces late binding and generates adjoints that will use the runtime types to determine what derivative computation to execute.

For example, \mintinline{python}{x * y} where \mintinline{python}{x} and \mintinline{python}{y} are scalars at runtime results in a scalar multiplication. However, if either of the two variables is a NumPy \mintinline{python}{ndarray} object, the multiplication operator is dispatched to perform broadcasting followed by element-wise multiplication. The adjoint of this operation requires summing over the broadcasted axes. Tangent will generate code that uses type checking to ensure that the correct adjoint calculation is performed based on the runtime types.

Similarly, the initialization and addition of gradients cannot be generated statically. We introduce \mintinline{python}{add_grad} and \mintinline{python}{init_grad} operators which use multiple dispatch. For example, \mintinline{python}{init_grad(x)} will return \mintinline{python}{0} if \mintinline{python}{x} is a scalar, but will return \mintinline{python}{numpy.zeros_like(x)} if \mintinline{python}{x} is an \mintinline{python}{ndarray}.

\subsection{Lazy evaluation}

A common performance bottleneck in the context of AD and array programming is that initializing the gradient of a large array results in allocating a large zero array. When gradients are accumulated later on this large array of zeros is added to a partial gradient, which is effectively a no-op. In general, the gradient initialization and addition might happen in different functions, making it non-trivial to statically optimize this case. To address this issue, Tangent lazily initializes gradients: Instead of allocating an array of zeros, Tangent returns a special \mintinline{python}{ZeroGrad} object. The \mintinline{python}{add_grad} operator uses multiple dispatch to return the other argument when either argument is of the type \mintinline{python}{ZeroGrad}.

 \subsection{Static optimizations}
 
 When constructing the adjoint of a function, some of the code of the forward pass might become dead code. The opportunity for removing unused code only grows when taking higher order derivatives. One of the advantages of SCT is that the resulting code can be optimized by an optimizing compiler whose dead code elimination (DCE) pass would address this problem. However, Python is an interpreted language, and very few optimizations are applied before its execution. For this reason, Tangent includes a small Python optimizing compiler toolchain which constructs a control-flow graph (CFG) on which forward dataflow analysis is performed. Tangent uses this to perform dead code elimination on generated adjoints. The same machinery is used to perform algebraic simplifications and constant propagation. Note that although these optimizations are hard to perform on Python in general, we can exploit the fact that Tangent operates on a more limited subset of Python which is more amenable to analysis (see Section~\ref{sec:limitations} for details).

   \begin{listing}[ht]
    \begin{minipage}[t]{0.55\textwidth}
    \begin{minted}[fontsize=\small]{python}
# Raw generated code
def dfdx(x, by=1.0):
    # Initialize the tape
    _stack = tangent.Stack()
    y = None
    # Beginning of forward pass
    tangent.push(_stack, y, '_19429e9f')
    y = x
    # Beginning of backward pass
    _y = y
    # Grad of: y = x
    y = tangent.pop(_stack, '_19429e9f')
    _bx = tangent.copy(by)
    by = tangent.init_grad(y)
    bx = _bx
    return bx
    \end{minted}
        \vspace{5pt}
    \end{minipage}
    \begin{minipage}[t]{0.4\textwidth}
    \begin{minted}[fontsize=\small]{python}
# Optimized generated code
def dfdx(x, by=1.0):
    y = x
    # Grad of: y = x
    _bx = tangent.copy(by)
    bx = _bx
    return bx
    \end{minted}
    \end{minipage}
       \caption{A simple example of Tangent's optimization capabilities as applied to the gradient function of \mintinline{python}{def f(x): y = x; return y}. Note that the original transformation includes the writing and reading of \mintinline{python}{y} to and from the tape, and contains dead code in initializing the gradient of \mintinline{python}{y} which is never returned. Tangent's dataflow analysis is able to match the tape reads and writes and understands that the value of \mintinline{python}{y} is the same, allowing it to aggressively optimize the function.}
       \label{lst:tapes}
       \end{listing}
 
  A central problem in reverse mode AD is that intermediate values are required to be kept alive after they go out of scope since they might be needed by their adjoint. For example, if a function contains \mintinline{python}{z = x * y} the variables \mintinline{python}{x} and \mintinline{python}{y} cannot be deleted after the function returns since the backward pass requires their values to calculate \mintinline{python}{dx = dz * y} and \mintinline{python}{dy = dz * x}. Tangent, like most SCT frameworks, uses a global stack (tape) to store intermediate variables on in order to ensure they are kept alive. Hence, before the function returns, \mintinline{python}{x} and \mintinline{python}{y} are pushed onto this stack and they will be popped off the stack right before the adjoint calculation. Note that the trace used in OO is also referred to as a tape, the difference being that the tape in OO stores not only the intermediate variables, but also the operations performed.
  
  In order to perform DCE effectively on the generated code, our dataflow analysis follows variables uses through their respective pushes (reads) and pops (writes) in the primal and adjoint code. This highlights the close interaction required between the optimizing compiler and the AD machinery for maximum performance. To enable the dataflow analysis to match reads and writes they are augmented in the source code with unique hashes (see Listing~\ref{lst:tapes}).
 
\subsection{Persistent data structures}\label{sec:arrays}

\begin{figure}
    \centering
    
\end{figure}

  \begin{listing}
    \begin{minipage}{0.5\textwidth}
    \includegraphics[width=\textwidth]{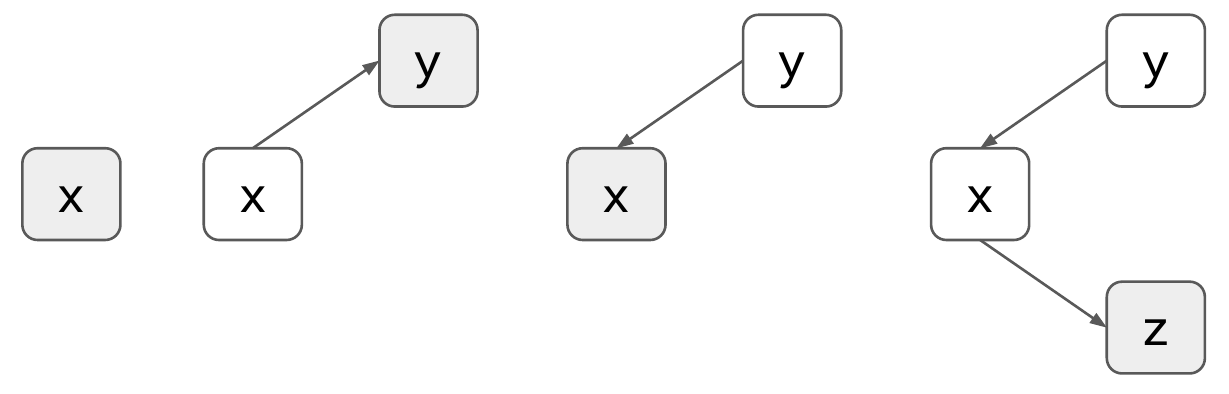}
\end{minipage}
    \begin{minipage}{0.5\textwidth}
        \begin{minted}[xleftmargin=0.1\textwidth,fontsize=\small]{python}
# Create handle to original version
x_copy = copy.copy(x)
# Create new node y
x[i] = v
# Restore original version
print(x_copy)
# Modify old version to create z
x_copy[i] = v
    \end{minted}
    \end{minipage}
\caption{Illustration of the persistent array data structure. Root nodes are gray and and edges represent deltas.}
    \label{lst:pa}
  \end{listing}
  
 AD is problematic in the context of mutability. If \mintinline{python}{x} and \mintinline{python}{y} from the previous example are mutable arrays, their value could have been changed by an in-place operation, resulting in an incorrect adjoint calculation. For this reason, arrays are in principle immutable in existing AD frameworks for ML such as TensorFlow, Autograd, and Theano. PyTorch allows users to mutate arrays if they can guarantee that the previous version will not be needed by the backward pass, otherwise an error will be thrown. This makes algorithms which rely on mutating arrays in place inefficient and difficult to express.
  
  Persistent data structures~\cite{driscoll1989making} are data structures that are effectively immutable: They are mutable data structures where all previous versions can be accessed. Unlike truly immutable data structures, different versions of persistent data structures may share memory and hence can be more memory-efficient, although accessing previous versions might require extra work. Functional languages often use persistent data structures for implementing, e.g., linked lists, trees, stacks. We note that persistent array data structures can be used to support mutability efficiently in the context of AD.\@
  
  By default, Tangent handles index assignments (\mintinline{python}{x[i] = y}) efficiently by copying only the affected subarray \mintinline{python}{x[i]} onto the tape. To deal with mutability in full generality Tangent also introduces a persistent array data structure with support for index assignment as well as inserting and deleting rows at the end. Each time the array is modified, the delta with the previous version is stored. Since previous versions can be modified as well, this results in a version tree where the root contains the current array in memory and other versions of the array are represented by leaf nodes (see Listing~\ref{lst:pa}). If the user attempts to read a specific version of the array, the deltas on the path from the leaf to the root of the version tree are applied in order to reconstruct the array in memory. When the handle to a specific array version is destroyed, the deltas are garbage collected. We note that in the context of reverse mode AD the most common access pattern is a linear series of mutations during the forward pass, followed by accessing the arrays in reverse order during the backward pass. In this case, our persistent array results in optimal memory and time complexity.
  
  As an example, consider the double loop from Listing~\ref{lst:persistent}, which is a simplification of a neural lattice language model from~\cite{buckman2018neural}. Given an outer loop with \(n\) iterations, an inner loop with \(m\) iterations, and a vector dimensionality of \(d\), the complexity of this algorithm is \(O(n^2md)\) for immutable arrays. When using regular NumPy arrays, Tangent will intelligently handle index assignments and only copy the affected subarray onto the tape, bringing the complexity down to \(O(n^2d + ndm)\). When a persistent array is used, the complexity goes down to \(O(ndm)\). When using persistent arrays, Tangent's runtime and memory complexity is determined only by the amount of data that is inserted, deleted or modified. In contrast, most libraries will have the gradient's runtime and memory complexity grow linearly with the number of times an array is modified. The technique described in~\cite{rae2016scaling} for memory-augmented networks is also a special case of using persistent arrays.
  
  \begin{listing}
    \begin{minipage}{0.6\textwidth}
  \begin{tikzpicture}
\begin{axis}[height=0.5\textwidth,width=\textwidth,xtick={100,300,...,1000},ylabel near ticks,ytick={0,2,...,8},xmin=100,xmax=1000,ymin=0,ymax=8,ylabel={Runtime (s)},xlabel={Outer loop length (iterations)}]
\addplot[black] table [x=n, y=t, col sep=comma] {immutable.csv};
\addlegendentry{Immutable arrays}
\addplot[black,dashed] table [x=n, y=t, col sep=comma] {persistent.csv};
\addlegendentry{Persistent array}
\addplot[black,densely dotted] table [x=n, y=t, col sep=comma] {index.csv};
\addlegendentry{Store subarray}
\end{axis}
\end{tikzpicture}
\end{minipage}
    \begin{minipage}{0.4\textwidth}
        \begin{minted}[xleftmargin=0\textwidth,fontsize=\small]{python}
def f(x, OUTER):
    r = numpy.zeros(DIM)
    for _ in range(OUTER):
        x = append(x, r)
        for _ in range(INNER):
            y = numpy.add(x[-1], 1.)
            x = setitem(x, -1, y)
    return numpy.mean(x)
    \end{minted}
    \end{minipage}
\caption{Runtime for a simplified version of a lattice language model with dimension 2000 and inner loop of 15 iterations. Results are an average of 10 runs.}
    \label{lst:persistent}
  \end{listing}

  \section{Limitations}
  
  \label{sec:limitations}

SCT relies on the ability to perform dataflow analysis to determine which variables are `active' i.e.\ which variables affect the output of the function whose derivative we are constructing. To this end, Tangent is restricted to a subset of Python where these analyses are feasible. Note that these restrictions only apply to statements involving active variables.

\begin{enumerate}
\item Functions that modify a variable in-place must also return that variable. Hence, \mintinline{python}{numpy.add(a, b, out=a)} is disallowed and should be written as \mintinline{python}{a = numpy.add(a, b)}. Likewise, a user-defined function that modifies \mintinline{python}{x} in-place using \mintinline{python}{x[i] = v}, must have \mintinline{python}{x} as a returned value.
\item Closures are not supported since closures with free variable references lead to a problem sometimes referred to as `perturbation confusion'~\cite{siskind2005perturbation}, which is non-trivial to address. Additionally, Python uses lexical, not dynamic scoping, so writing adjoint values into the same scope where primal values are read is not straightforward. 
\item Object methods are not currently supported because it is non-obvious what the partial derivative with respect to a member variable is.
\item In order to perform AD, the function and its source code must be resolvable at the time that the AD transformation is applied. This means that higher-order and nested functions are not supported. Tangent could apply additional AD passes at runtime to avoid this limitation.
\item Some Python syntax is not (yet) supported e.g.\ \mintinline{python}{try} and \mintinline{python}{except} statements, as well as \mintinline{python}{break} and \mintinline{python}{continue}.
\end{enumerate}

If the return value of a function is not used, it is assumed that its inputs were unchanged. This allows statements such as \mintinline{python}{print(numpy.mean(x))} to be used without interfering with the AD transformation.
  
  \section{Performance}
  
  Tangent was not designed with raw performance in mind. Instead, it intends to strike a balance between usability and good software design practices, while exploring the feasibility and implementation details of applying SCT to dynamically typed languages. That said, Tangent's lack of runtime overhead combined with static optimizations and lazy gradient initialization means that its runtime performance is competitive with existing frameworks (see Listing~\ref{lst:benchmark}).

 \begin{listing}
    \begin{minipage}{0.45\textwidth}
  \begin{tikzpicture}
\begin{axis}[ymode=log,xmode=log,log basis x={2},legend pos=north west,height=1.2\textwidth,width=\textwidth,ylabel={Seconds per batch (s\textsuperscript{-1}}),xtick=data,xmin=64,xmax=8192,xlabel={Model size},ylabel near ticks]
\addplot[black] table [x=n, y=t, col sep=comma] {tangent.csv};
\addlegendentry{Tangent}
\addplot[black,dashed] table [x=n, y=t, col sep=comma] {autograd.csv};
\addlegendentry{Autograd}
\addplot[black,densely dotted] table [x=n, y=t, col sep=comma] {tensorflow.csv};
\addlegendentry{TensorFlow}
\end{axis}
\end{tikzpicture}
\end{minipage}
    \begin{minipage}{0.55\textwidth}
        \begin{minted}[xleftmargin=0.05\textwidth,fontsize=\small]{python}
def logsumexp(x):
    return numpy.log(numpy.sum(numpy.exp(x), 
                     axis=-1, keepdims=True))

def logsoftmax(logits):
    return logits - logsumexp(logits)

def softmax_xent(logits, y):
    return -numpy.sum(
        logsoftmax(logits) * y, axis=-1)

def mlp(x, w1, b1, wout, bout, label):
    h1 = numpy.tanh(numpy.dot(x, w1) + b1)
    out = numpy.dot(h1, wout) + bout
    loss = numpy.mean(softmax_xent(out,label))
    return loss

autograd_dmlp = autograd.multigrad(
    mlp, argnums=(1, 2, 3, 4))
tangent_dmlp = tangent.grad(
    mlp, wrt=(1, 2, 3, 4))
    \end{minted}
    \end{minipage}
\caption{Runtime for a simple MLP. We vary the input size and hidden layer size, which are set to the same value. The reported runtime is averaged over 50 runs with a batch size of 16. Run on a Xeon E5-1650 v3 @ 3.5 GHz, 64GB of RAM, with Ubuntu 14.04 on Python 2.7 with MKL.}
    \label{lst:benchmark}
  \end{listing}
  
  \section{Conclusion}
  
  In this work we introduced the AD library Tangent. Tangent is the first application of source-code transformation on a dynamically typed language such as Python. It uses several novel approaches, such as persistent data structures and lazy evaluation to ensure good performance. Machine learning models are natural and easy to express and debug in Tangent using many features that are not available in other frameworks e.g.\ mutable arrays, inspectable derivative code, and modifying gradients by injecting arbitrary code in the backward pass.
  
  We believe Tangent is an important step on the path to fully general differentiable programming. Instead of an ML-framework, Tangent can be seen as the addition of the gradient operator to the Python language, without the need for metaprogramming or separate derivative interpreters (OO). This means that the user can write normal Python code while the entire Python ecosystem including debuggers, profilers, and introspection capabilities, become part of the ML toolkit. This allows users to express models more naturally and debug them more easily.

\bibliographystyle{plainnat}
\bibliography{tangent}

\end{document}